%% file: main.tex
\documentclass{article}

\usepackage[nonatbib,preprint]{neurips_2022}

\usepackage[square,sort,comma,numbers]{natbib}
\usepackage[utf8]{inputenc} 
\usepackage[T1]{fontenc}    
\usepackage[dvipsnames]{xcolor}         
\usepackage[pagebackref=true,breaklinks=true,letterpaper=true,colorlinks,bookmarks=false, citecolor=ForestGreen]{hyperref}
\usepackage{url}            
\usepackage{booktabs}       
\usepackage{amsfonts}       
\usepackage{nicefrac}       
\usepackage{microtype}      

\input{lib/my_instructions.tex}

\title{LightViT: Towards Light-Weight Convolution-Free Vision Transformers}

\author{%
    Tao Huang${}^{1,2}$ \quad Lang Huang$^3$ \quad Shan You$^{1}$\thanks{Correspondence to: Shan You $<$\texttt{youshan@sensetime.com}$>$.} \quad Fei Wang$^4$ \quad Chen Qian$^1$ \quad Chang Xu$^2$\\
	$^1$SenseTime Research \\
	$^2$School of Computer Science, Faculty of Engineering, The University of Sydney\\
	$^3$The University of Tokyo \quad
	$^4$University of Science and Technology of China \\
}

\begin{document}

\maketitle

\vspace{-4mm}
\begin{figure}[h]
    \centering
    \begin{minipage}{0.4\linewidth}
        \centering
        \includegraphics[width=1\linewidth]{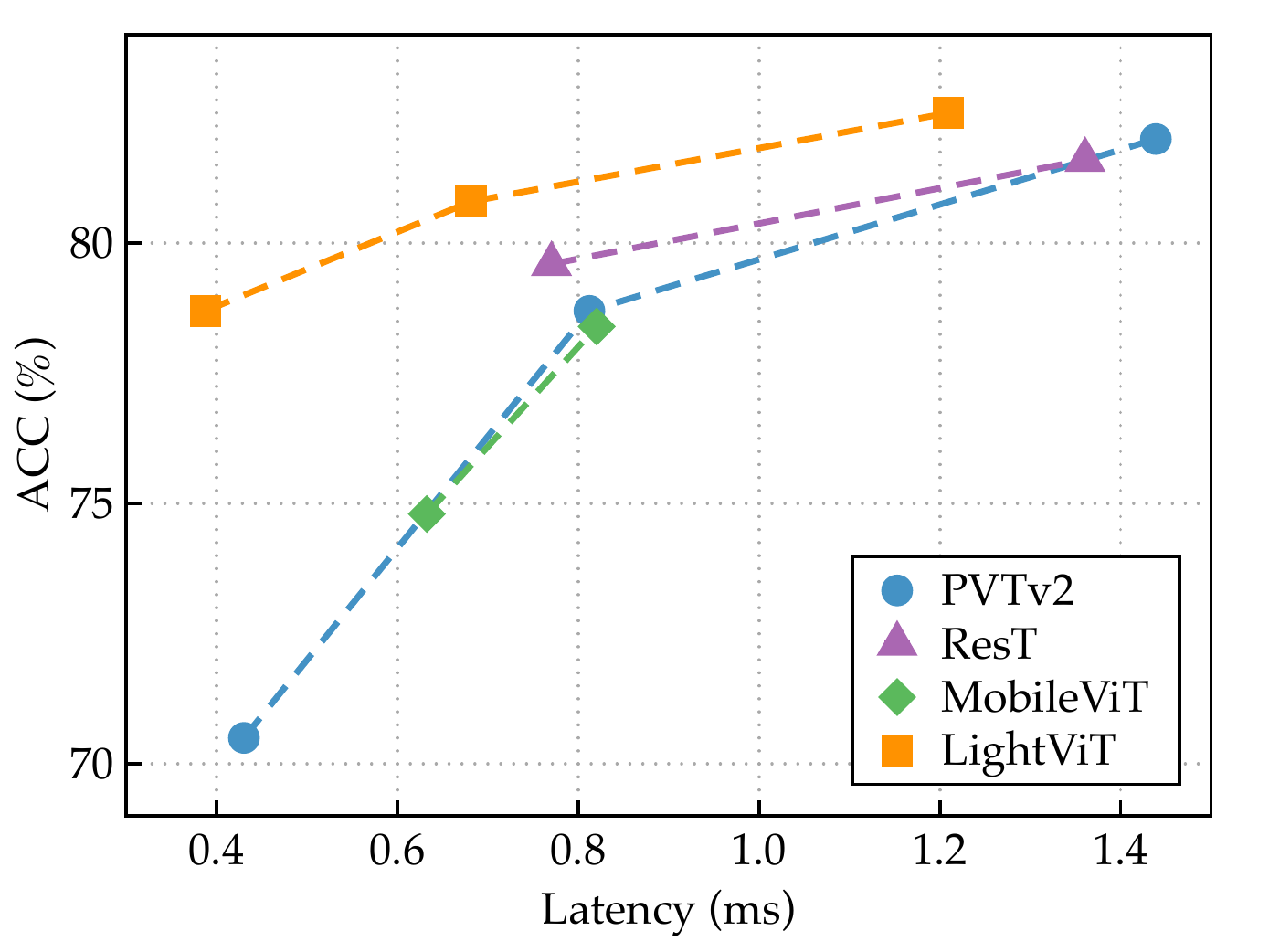}
    \end{minipage}
    \begin{minipage}{0.4\linewidth}
        \centering
        \includegraphics[width=1\linewidth]{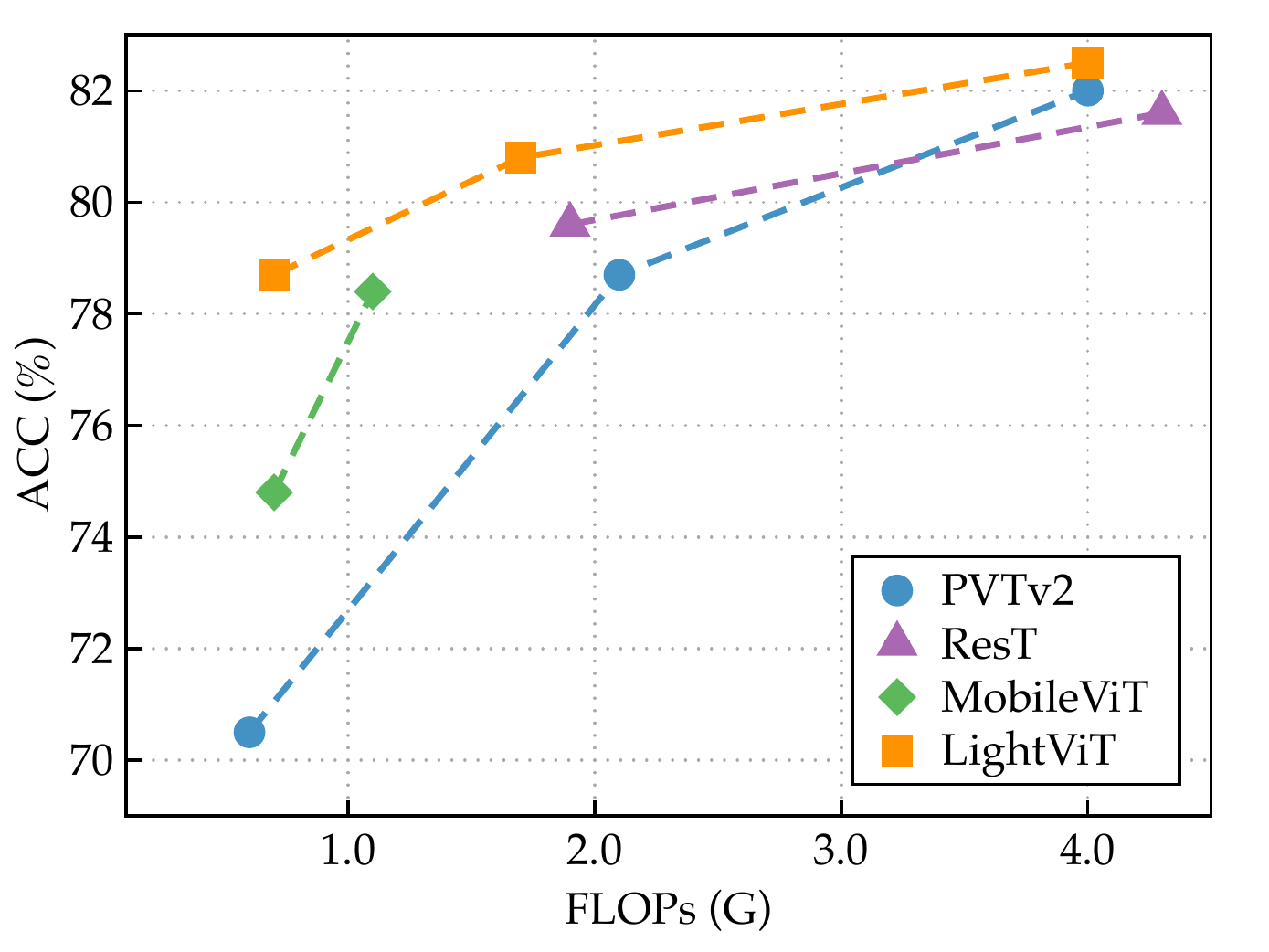}
    \end{minipage}
    \vspace{-2mm}
    \caption{Comparisons of the proposed LightViT and other efficient ViTs on ImageNet.}
    \label{fig:f1}
\end{figure}

\begin{abstract}
Vision transformers (ViTs) are usually considered to be less light-weight than convolutional neural networks (CNNs) due to the lack of inductive bias. Recent works thus resort to convolutions as a plug-and-play module and embed them in various ViT counterparts. In this paper, we argue that the convolutional kernels perform information aggregation to connect all tokens; however, they would be actually unnecessary for light-weight ViTs if this explicit aggregation could function in a more homogeneous way. Inspired by this, we present LightViT as a new family of light-weight ViTs to achieve better accuracy-efficiency balance upon the pure transformer blocks without convolution. Concretely, we introduce a global yet efficient aggregation scheme into both self-attention and feed-forward network (FFN) of ViTs, where additional learnable tokens are introduced to capture global dependencies; and bi-dimensional channel and spatial attentions are imposed over token embeddings. Experiments show that our model achieves significant improvements on image classification, object detection, and semantic segmentation tasks. For example, our LightViT-T achieves 78.7\% accuracy on ImageNet with only 0.7G FLOPs, outperforming PVTv2-B0 by 8.2\% while 11\% faster on GPU. Code is available at \href{https://github.com/hunto/LightViT}{https://github.com/hunto/LightViT}.
\end{abstract}

\section{Introduction}

Recently, vision transformers (ViTs) have gained noticeable success on vision tasks such as image classification \cite{dosovitskiy2020image, touvron2021training, liu2021swin, su2021vision}, object detection \cite{liu2021swin, li2022exploring}, and semantic segmentation \cite{strudel2021segmenter, xie2021segformer}. However, despite the state-of-the-art performance in large-size ViT models, their light-weight counterpart will lose its advantage over the typical convolutional neural networks (CNNs). For instance, it is observed that DeiT-Ti~\cite{touvron2021training} and PVTv2-B0~\cite{wang2022pvt} can achieve 72.2\% and 70.5\% accuracies on ImageNet, while the typical CNN model RegNetY-800M \cite{radosavovic2020designing} achieves 76.3\% accuracy with similar FLOPs, which seems a catastrophic failure for ViTs in terms of light-weight models. 

It is recognized that CNNs are generally more efficient for their intrinsically-biased architecture designs, such as parameter sharing, local information aggression, and spatial reduction. Therefore, to enhance the light-weight property of ViTs, recent works mainly borrow the inductive bias from CNNs to develop various counterparts in a hybrid or heterogeneous manner, \ie~integrating convolutions into transformer blocks as a plug-and-play module. For example,
ResT~\cite{zhang2021rest} proposes to leverage convolutions to reduce the spatial dimensions of key and values in self-attention; LVT~\cite{yang2021lite} adopts convolutions to perform local self-attention for low-level features and multi-scale attentions for high-level features. Moreover, some methods \cite{wu2020visual, srinivas2021bottleneck, mehta2021mobilevit} aim to improve CNNs via interpreting self-attentions into existing CNN blocks. A recent study MobileViT~\cite{mehta2021mobilevit} incorporates transformer into MobileNetV2~\cite{sandler2018mobilenetv2} to obtain global representations in the upper stages.

So far, the community shows that convolution seems to be essential for efficient ViTs. However, \textit{is convolution really necessary for light-weight ViTs?} \textit{Can't we have an efficient homogeneous ViT with no convolution but only the transformer block?} In this paper, we get down to investigating this problem and hope to push the limit of light-weight ViT one step further. By rewinding the convolution in hybrid ViTs, we regard it as a way of information aggregation since it builds explicit connections to all the tokens through shared kernels. In this way, we are inspired to introduce these aggregation priors to the ViTs as well, which stimulates new design for the two key components in transformer blocks, \ie, self-attention and feed-forward network (FFN):

\begin{itemize}
    \item For self-attention, we leverage the local window attention~\cite{liu2021swin} for effective spatial priors and efficient calculation. In particular, we
    propose to introduce learnable \textit{global tokens} to aggregate the information of local tokens by modeling their global dependencies. Then these global dependencies are broadcast into every local token. In this way, each image token could be more informative since it benefits from both local and global features as Figure \ref{fig:global_attn} (a). Note that the global dependencies can be calculated quite efficiently. 
    \item For FFN, as the only non-linearity in plain transformer block, it plays an important role in feature extractions by modeling feature patterns and implicitly capturing the spatial dependencies.
    However, its representation power would be restricted due to the small channel dimensions in light-weight models. Therefore, we propose a bi-dimensional attention module to 
    explicitly aggregate the global dependencies among spatial and channel dimensions, thus the capacity would be enhanced since features will be filtered more adaptively. 
    
\end{itemize}

Based on the new self-attention and FFN, we also make empirical studies to give a more practical design for efficient ViTs, which help us achieve a better efficiency-accuracy tradeoff. For example, we observe that the early stages in hierarchical ViTs are inefficient due to a large number of tokens in self-attention, and thus propose to build ViT stages from a moderate dimension ($\mathrm{stride}=8$) such as discarding the Stage 0 as Figure \ref{fig:arch}. As a result, we can develop a new family of light-weight convolution-free ViTs dubbed \textit{LightViT}. Extensive experiments show that our LightViT does enjoy significant superiority and efficiency advantage over various computer vision benchmarks. For example, as shown in Figure~\ref{fig:f1}, our LightViT-S achieves 80.8\% accuracy on ImageNet, significantly outperforms ResT-Small~\cite{zhang2021rest} by 1.2\% with 0.2G smaller FLOPs and 14\% faster in inference.

\section{Related Work}

\subsection{Efficient vision transformers} 
Recent approaches~\cite{wang2021pyramid, zhang2021rest, wang2022pvt} on efficient ViTs mainly focus on interpreting convolutions into transformer blocks. PVT~\cite{wang2021pyramid} conducts a CNN-like hierarchical structure and adopts convolutions to reduce the spatial dimensions of self-attention and perform feature downsampling. PVTv2~\cite{wang2022pvt} further improves PVT by introducing overlapping patch embedding and convolutional feed-forward network. ResT~\cite{zhang2021rest} proposes a memory-efficient self-attention by compressing the spatial dimensions and projects the interaction across the attention-heads dimension using convolutions. LVT~\cite{yang2021lite} introduces convolutions in self-attention to perform local self-attention for low-level features and multi-scale attention for high-level features. Different from previous models built upon the ViT structure, MobileViT~\cite{mehta2021mobilevit} aims to improve mobile CNNs by incorporating attention into MobileNetV2 blocks for better global representations. This paper investigates a new variant of efficient ViTs without using convolutions in blocks.

\subsection{Window-based vision transformers}
Despite the success of plain ViTs on image classification, it remains challenging for downstream tasks since the computation cost would grow quadratic to the image size on these high-resolution tasks. Recent works~\cite{wang2021pyramid, wang2022pvt, liu2021swin, chu2021twins} conduct hierarchical structures with multiple stages like CNNs to make ViTs more efficient and friendly to existing frameworks. Among these methods, window-based methods~\cite{liu2021swin, chu2021twins} adopt local window attention to partition image tokens into multiple non-overlapped windows and perform self-attentions inside each window, yielding a linear computation complexity to the image size. 

However, local window attention has been observed to have limited receptive fields and weak long-range dependencies. Therefore, some methods propose to bring global interactions to the local window attention. Twins~\cite{chu2021twins} applies global attention to image tokens (queries) and window representations (keys and values) summarized by convolutions. MSG-Transformer~\cite{fang2021msg} binds learnable message tokens on each local window and adopts channel shuffle among these tokens to exchange information. Focal transformer~\cite{yang2021focal} performs local window attentions using keys and values down-sampled with different strides, thus aggregating information on multiple receptive fields. Nevertheless, these global information aggregations still have a quadratic computation cost to the input image size and encounter large computation cost especially on large resolutions. In this paper, we introduce global tokens to aggregate the global information freely on the whole feature map, which only has linear computation complexity to the input image size and brings noticeable improvements with negligible FLOPs increment.

\begin{figure}[t]
    \centering
    \hspace{-4mm}
    \subfloat[\textbf{Local-global broadcast of attention.}] {
        \centering
        \begin{minipage}[c][0.3\linewidth]{0.7\linewidth}
            \centering
            \includegraphics[width=0.9\linewidth]{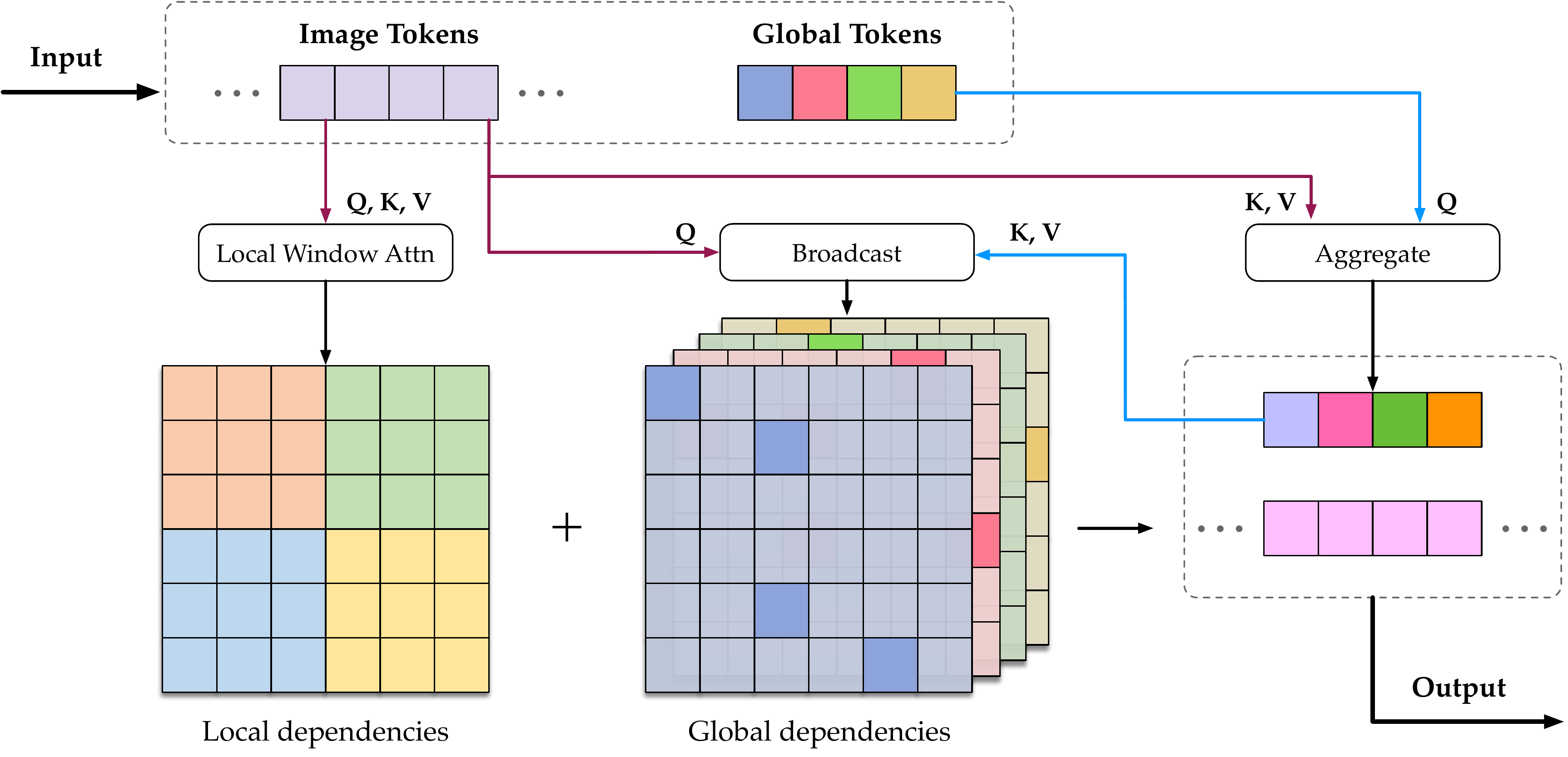}
        \end{minipage}
    }
    \hspace{4mm}
    \subfloat[\textbf{Bi-dimensional attn.}] {
        \centering
        \begin{minipage}[c][0.3\linewidth]{0.25\linewidth}
            \centering
            \includegraphics[width=1.\linewidth]{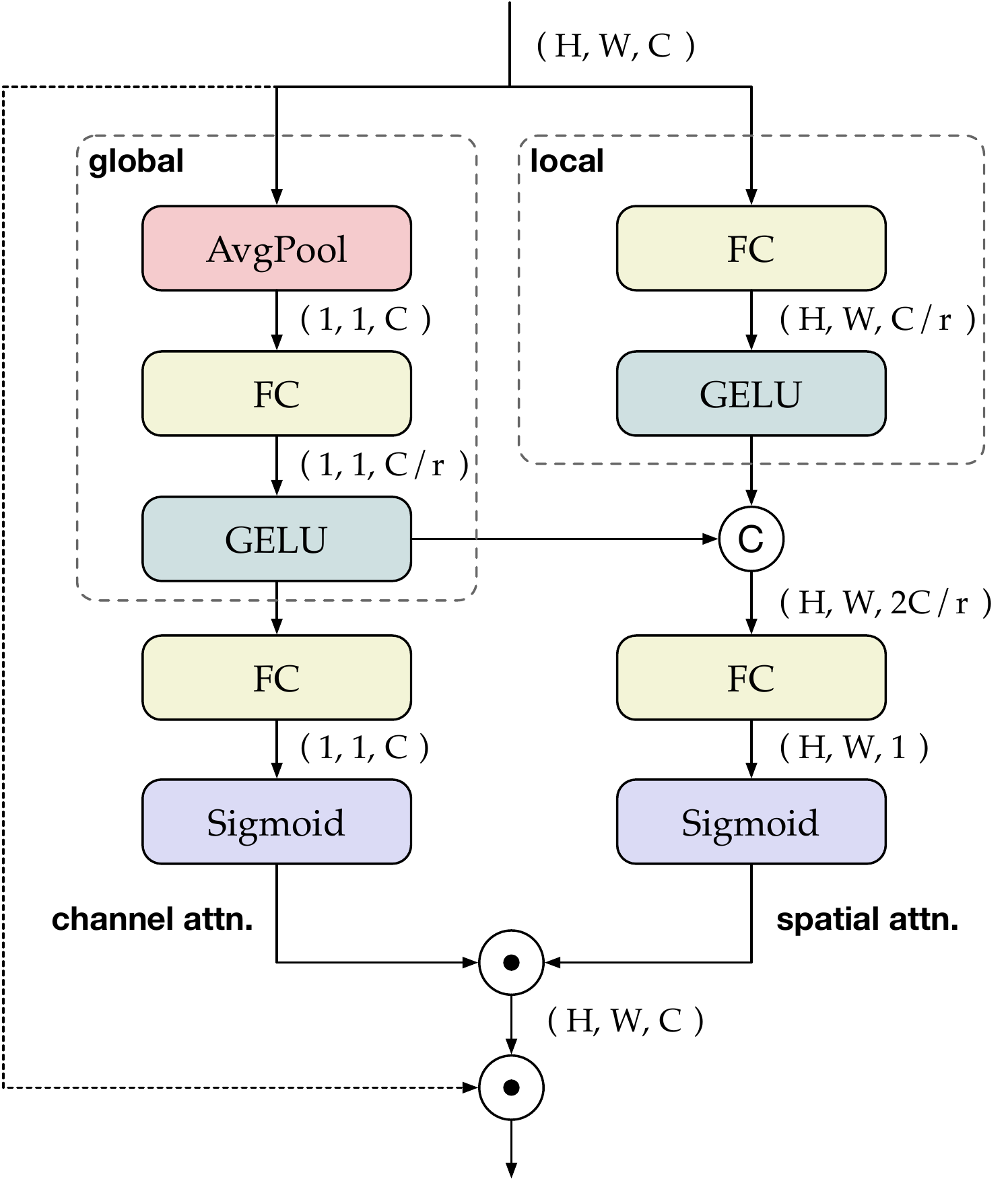}
        \end{minipage}
    }
    \caption{\textbf{Our proposed efficient feature aggregation on attention and FFN.} (a) Besides local window self-attention (window size is $3$ here), we propose local-global broadcast of attention to broadcast the global information using additional global tokens. (b) The architecture of our bi-dimensional attention on FFN.}
    \label{fig:global_attn}
    \vspace{-2mm}
\end{figure}

\begin{figure}[t]
    \centering
    \includegraphics[width=1.\textwidth]{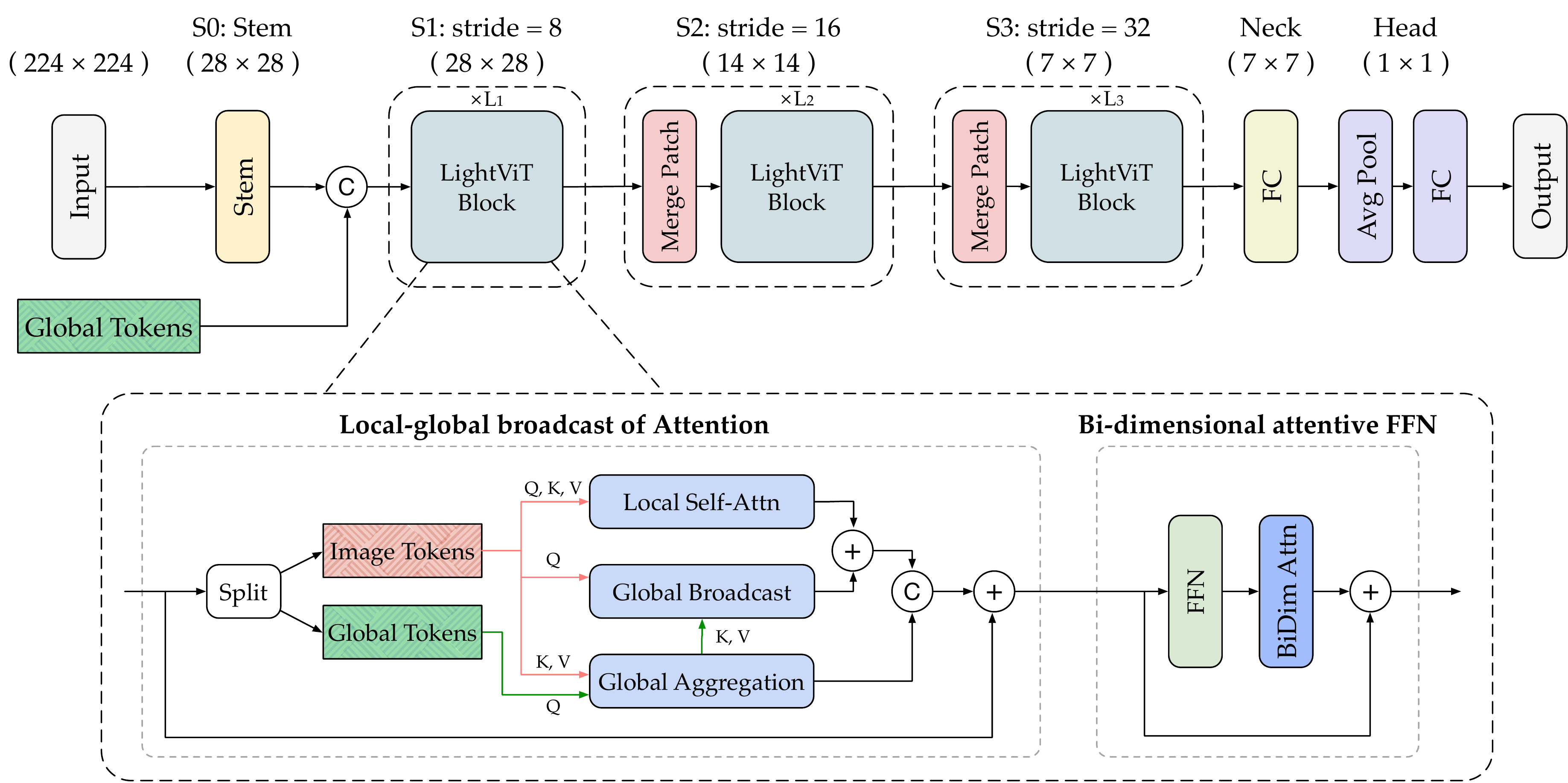}
    \vspace{-2mm}
    \caption{\textbf{LightViT architecture.} We follow a hierarchical structure design, but remove the $\text{stride}=4$ stage for better efficiency. \textcircled{c} denotes concatenation in the token axis, and \textcircled{+} denotes element-wise addition.}
    \label{fig:arch}
    \vspace{-2mm}
\end{figure}

\section{Efficient feature aggregation for LightViT}

In this section, we formally illustrate the two key designs for our LightViT, namely aggregated self-attention and FFN, which leverage local-global broadcast of attention and bi-dimensional attention, respectively. 

\subsection{Aggregated self-attention with local-global broadcast}

Self-attention among all the tokens is one of the key advantages of ViTs compared to local convolution. However, directly applying self-attention to the whole image requires quadratic computation complexity to the input image size. To reduce the computation cost, the typical local window self-attention~\cite{liu2021swin} partitions the feature map into multiple non-overlapping windows, then performs self-attention independently in each window. This paper utilizes the local window self-attention as the base module.

\textbf{Local self-attention.} Given an input feature map $\bm{X}\in\mathbb{R}^{H\times W \times C}$, instead of calculating attention on the flattened $H\times W$ patches, we partition $\bm{X}$ into non-overlapping windows with shape $(\frac{H}{S}\times \frac{W}{S}, S\times S, C)$, where $S$ denotes window size (we use $S=7$ following Swin~\cite{liu2021swin}), then apply self-attention within each local window, which is equivalent to the local window attention in Swin and Twins. Formally, the local self-attention is computed as
\begin{equation}
    \bm{X}_\textrm{local} = \mathrm{Attention}(\bm{X}_q, \bm{X}_k, \bm{X}_v) := \mathrm{SoftMax}(\bm{X}_q\bm{X}_k^\top)\bm{X}_v, 
\end{equation}
where $\bm{X}_q$, $\bm{X}_k$, and $\bm{X}_v$ are produced by Q, K, and V projections, respectively. As a result, the computation complexity $(H\times W)^2$ of self-attention is reduced to $(\frac{H}{S}\times \frac{W}{S})\times (S\times S)^2 = H\times W\times S\times S$.

The local self-attention can be an efficient and effective way to aggregate local dependencies with window priors. However, it has a drawback of lacking long-range dependencies and large receptive fields. For a light way to obtain global interactions, this paper proposes to first gather the valuable global dependencies to a small feature space, then broadcast the aggregated global information to the local features. This light information squeeze-and-expand scheme can enhance the local features with negligible computation cost, and we find it sufficient and effective in experiments.

\textbf{Global aggregation.} To gather global information in $\bm{X}$, we propose a learnable embedding $\bm{G}\in \mathbb{R}^{T\times C}$, which is computed along with image tokens in all LightViT blocks. The proposed embedding $\bm{G}$, dubbed as \textit{global token}, has two functions: global information aggregation and broadcast. As illustrated in Figure~\ref{fig:global_attn} (a), it first aggregates the global representations on the whole image feature map, then broadcasts the global information into the feature maps. All the information exchanges are performed in a homogeneous way using attention. Specifically, along with the computation of local self-attention, we gather the global representations using input global tokens $\bm{G}$ (queries) and image tokens $\bm{X}$ (keys and values), \ie,
\begin{equation}
    \hat{\bm{G}} = \mathrm{Attention}(\bm{G}_q, \bm{X}_k, \bm{X}_v) ,
\end{equation}
the output new tokens $\hat{\bm{G}}$ are then used in global broadcast and passed to next block for usage.

\textbf{Global broadcast.} With the aggregated global information, the aim is to broadcast it back to the image tokens, thus the image features can be enhanced by receiving global dependencies from tokens outside the local window. We perform this broadcast by adopting global tokens $\hat{\bm{G}}$ as keys and values in attention:
\begin{equation}
    \bm{X}_{\mathrm{global}} = \mathrm{Attention}(\bm{X}_q, \hat{\bm{G}}_k, \hat{\bm{G}}_v) ,
\end{equation}
then the final output image tokens are computed through element-wise addition on local and global features, \ie,
\begin{equation}
    \bm{X}_\mathrm{new} = \bm{X}_\mathrm{local} + \bm{X}_\mathrm{global}.
\end{equation}

Note that the computation complexities of global aggregation and global broadcast ($H\times W \times T$) are negligible, as the number of global tokens $T$ (\eg, $T=8$ in LightViT-T) is much smaller than image size $H \times W$ and window size $S\times S$ in LightViT.

\begin{figure}[t]
    \centering
    \includegraphics[width=1.\textwidth]{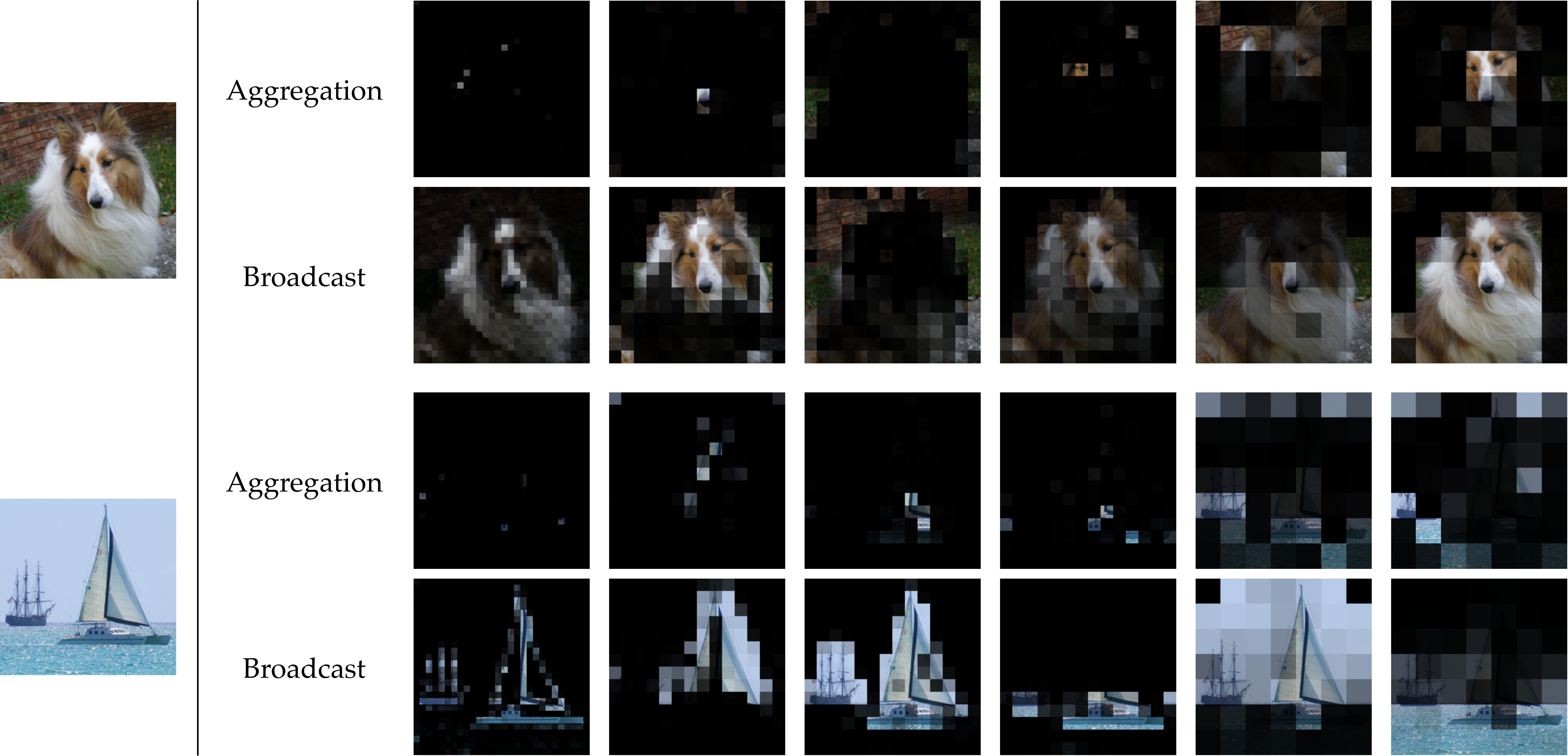}
    \vspace{-2mm}
    \caption{\textbf{Visualization of the learned global attentions of LightViT-T on ImageNet.} Our global tokens \textit{aggregate} key parts (\eg, nose and eyes of the dog) of the image (1st row), then \textit{broadcast} the information to related pixels globally (2nd row) to enhance and highlight the target objects.}
    \label{fig:vis_attn}
    \vspace{-2mm}
\end{figure}

We visualize the learned global attentions in Figure~\ref{fig:vis_attn}. We can see that, the global tokens first aggregate key information (\eg, nose and eyes of the dog) of the feature map through our global aggregation, then deliver the information to related pixels using global broadcast, and thus the features of target objects (\eg, dog and boat) can be enhanced and highlighted with global information.

\subsection{Aggregated FFN with bi-dimensional attention}

As the only non-linear part in transformer block, feed-forward network (FFN) plays an important role in feature extraction. Since all the tokens are forwarded point-wisely and share the same linear layers in FFN, the non-linear activations are usually conducted on enlarged channel dimensions produced by a linear layer for an effect and sufficient capture of feature patterns. However, the dimensions of channels are still insufficient in light-weight models, where the channels are limited to small ones for reducing the computation cost, and thus their performance is severely restricted. Another drawback of the plain FFN is the lack of explicitly dependency modeling on the spatial level, which is highly important to vision tasks. Though the spatial feature aggregations can be performed implicitly through the weight sharing among tokens, it is still challenging for the light ViTs to capture these implications. To this end, some ViT variants~\cite{li2021localvit, yuan2021incorporating, wang2022pvt} propose to aggregate spatial representations before the activation layer using convolutions, resulting in a noticeable increase in computation cost.

In this paper, inspired by the attention mechanisms~\cite{hu2018squeeze, woo2018cbam}, which are widely adopted to explicitly model the feature relationships in light-weight CNNs~\cite{tan2019efficientnet, howard2019searching}, we propose a bi-dimensional attention module to capture the spatial and channel dependencies and refine the features. As shown in Figure~\ref{fig:global_attn} (b), the module consists of two branches: channel attention branch and spatial attention branch. The channel attention branch first averages the input features on spatial dimension to aggregate global representations, which are then used to compute the channel attention with a linear transformation. For spatial attention, we model the pixel-wise relations by concatenating the global representations to every token features (local representations). To reduce the FLOPs, we add a linear reduction layer before the attention fully-connected (FC) layers following SE~\cite{hu2018squeeze}, and set the reduction ratio $r$ to $4$ in our models.

Our proposed bi-dimensional attention module can be used as a plug-and-play module for existing ViT variants. With only a slight increase in computation cost, it can explicitly model the spatial and channel relationships and enhance the representation power of FFN.

\begin{figure}
    \centering
    \subfloat[\textbf{Throughput of each stage.}] {
        \centering
        \begin{minipage}[c][0.25\linewidth]{0.35\linewidth}
            \centering
            \includegraphics[width=1.\linewidth]{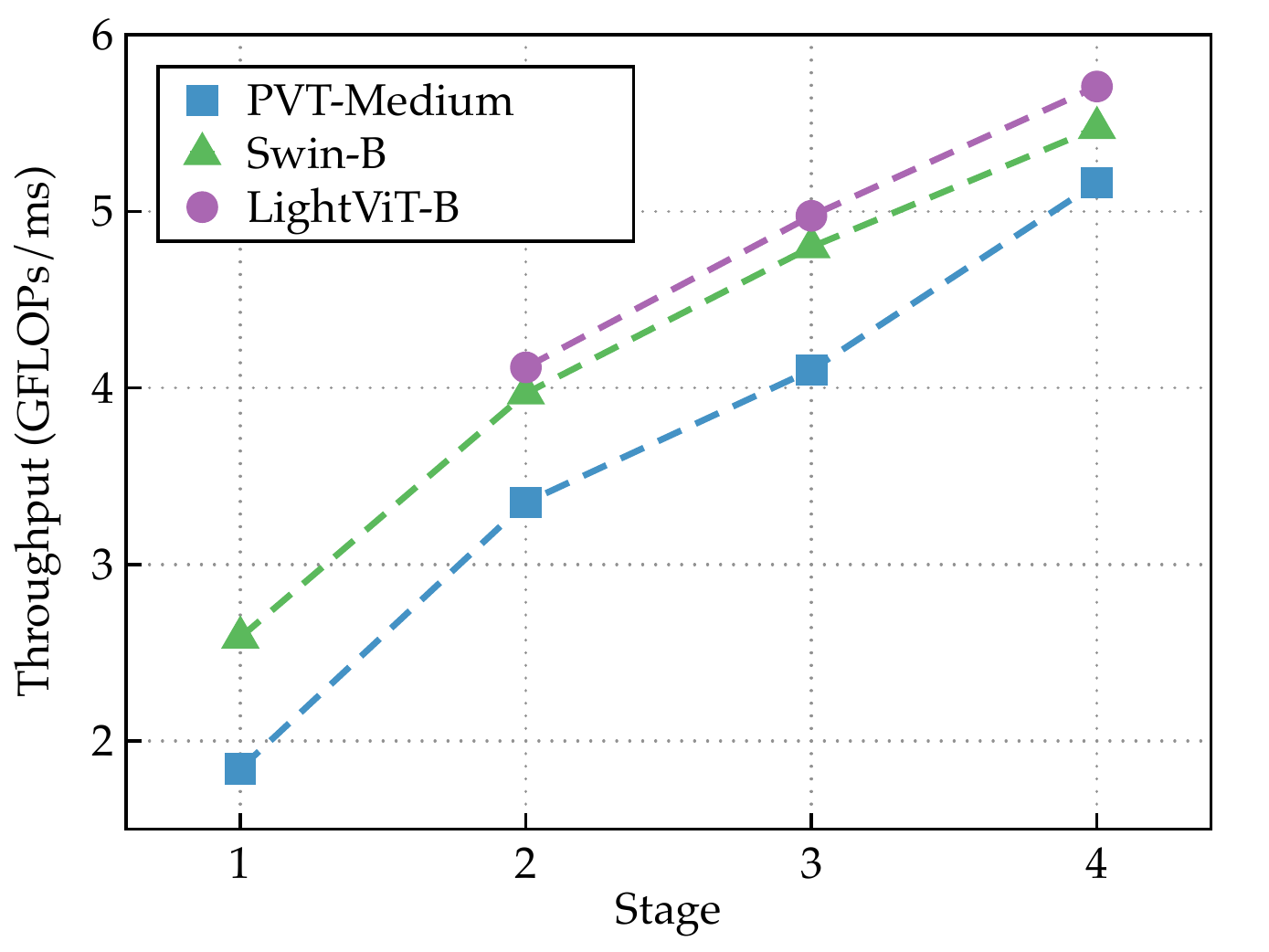}
        \end{minipage}
    }
    \subfloat[\textbf{FPN on 3 stages.}] {
        \centering
        \begin{minipage}[c][0.25\linewidth]{0.3\linewidth}
            \centering
            \includegraphics[width=1\linewidth]{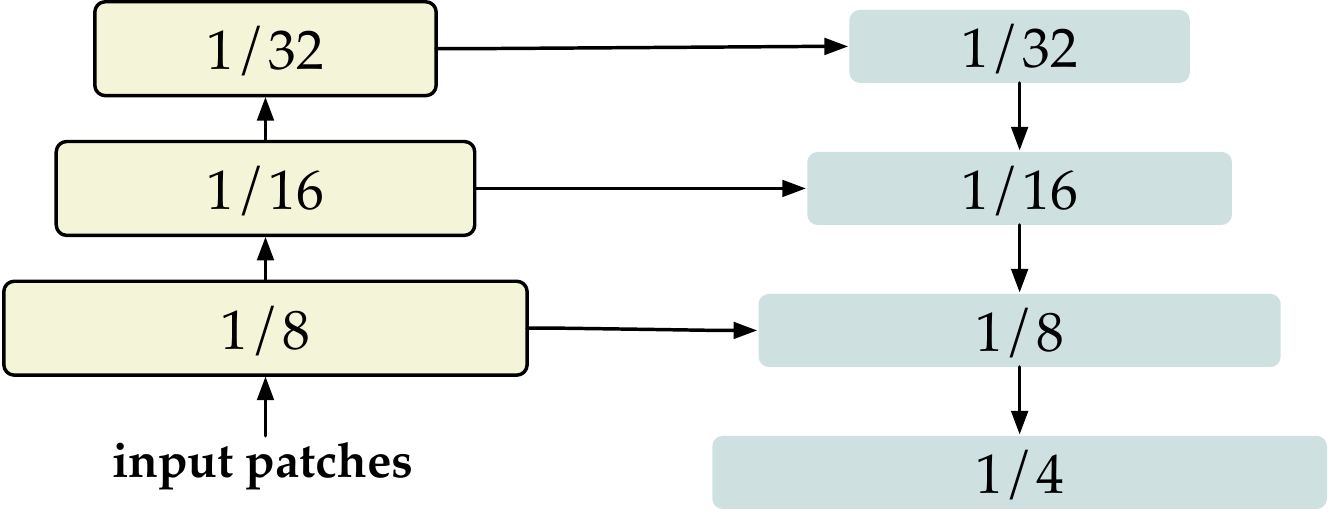}
        \end{minipage}
    }
    \subfloat[\textbf{Residual patch merging.}] {
        \centering
        \begin{minipage}[c][0.25\linewidth]{0.3\linewidth}
            \centering
            \includegraphics[width=1.05\linewidth]{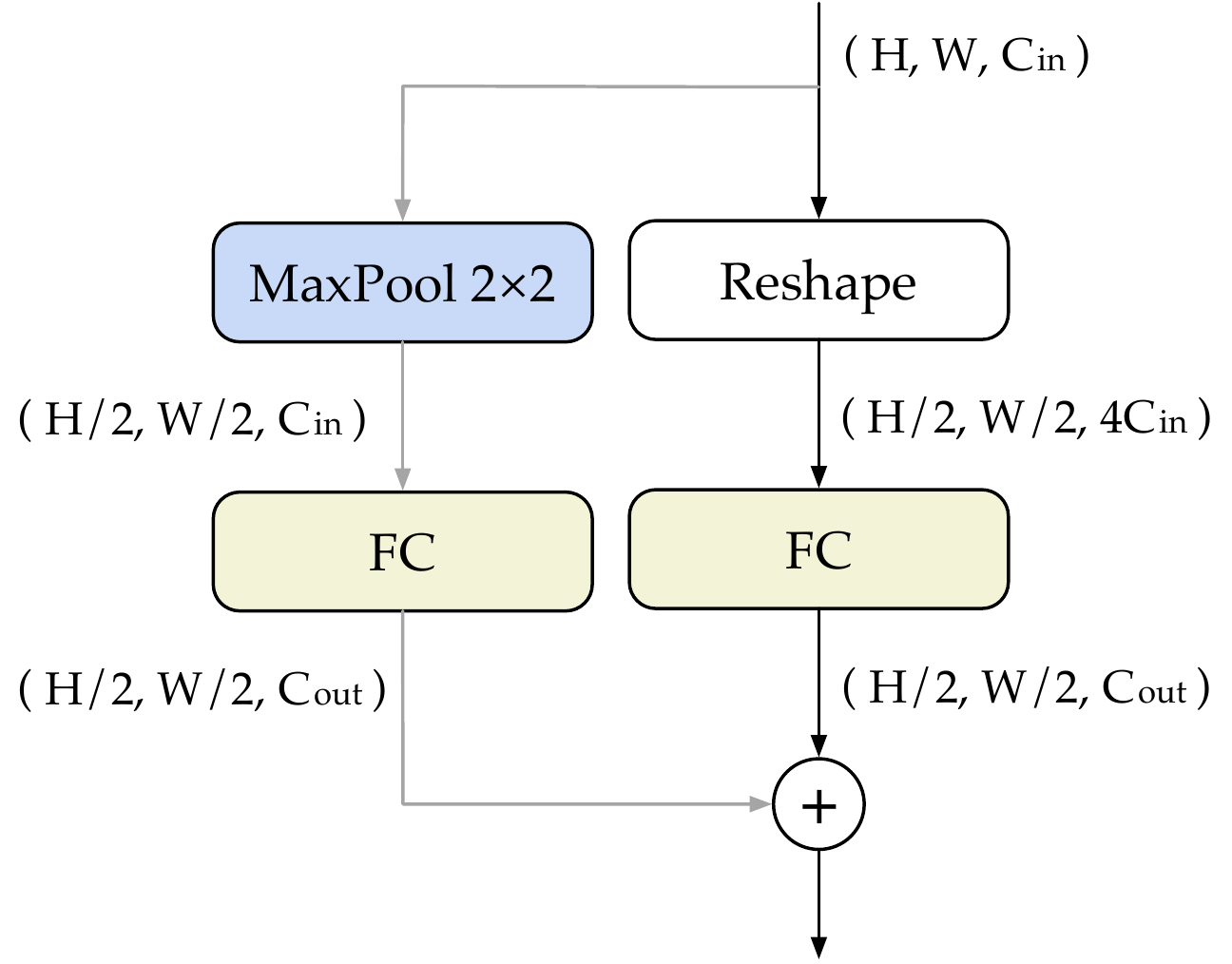}
        \end{minipage}
    }
    
\caption{(a): Throughput of each stage shows that the earlier stages are less efficient. (b): Architecture of our feature pyramid on object detection. (c): We extend the patch merging module in Swin~\cite{liu2021swin} with a cheap residual branch.}
\label{fig:design_choices}
\end{figure}

\nocite{zheng2021weakly, zheng2021ressl, you2020greedynas, huang2022dyrep, huang2022greedynasv2, su2021prioritized}

\section{Practical design for more efficient LightViT}
In this section, we formulate our design choices for LightViT through experiments. We empirically find that several improvements on the model components can lead to better performance and efficiency, and thus make our LightViTs more efficient. For fair comparisons, we keep the same FLOPs by adjusting the channels uniformly in experiments.

\subsection{Hierarchical structures with fewer stages}

Vision transformers with hierarchical structures~\cite{liu2021swin, wang2021pyramid} have shown great performance on image classification and downstream tasks. However, these methods have smaller inference speeds compared to the vanilla ViT. For example, the vanilla ViT model DeiT-S has a throughput of $961$ on GPU with 4.6G FLOPs, while PVTv2-B2 only has $695$ with 4.0G FLOPs. One major reason is that the earlier stages in hierarchical structures have larger numbers of tokens, making self-attention less efficient. As shown in Figure~\ref{fig:design_choices} (a), we measure the inference efficiency (speed / FLOPs) of hierarchical ViTs, and find that the earlier stages are less FLOPs-efficient than the later stages. As a result, in this paper, we remove the first $\mathrm{stride} = 4$ stage and keep the later $\mathrm{stride} = \{8, 16, 32\}$ stages in the hierarchical structures. The experimental results on ImageNet and COCO detection in the following table show that, our LightViT-T achieves significant efficiency improvement by removing the first stage, and even achieves higher accuracy.

\begin{table}[h]
    \vspace{-2mm}
	\renewcommand\arraystretch{1.3}
	\setlength\tabcolsep{1.2mm}
	\centering
	\footnotesize
	\begin{tabular}{l|cccc|c}
	    Method & Params (M) & FLOPs (G) & Throughput (images/s) & Top-1 (\%) & COCO det mAP \\
	    \shline
	    w/ 4 stages & 9.2 & 0.73 & 1643 & 77.9 & 37.5\\
	    LightViT-T & 9.4 & 0.73 & 2578 & \textbf{78.7} & \textbf{37.8}\\
	\end{tabular}
	\vspace{-2mm}
\end{table}

On downstream tasks such as object detection, it usually adopts a 4-stage feature pyramid network (FPN) \cite{lin2017feature}. Removing the first stage may have a possible risk of weakening the transfer performance. In this paper, we show that directly adopting a 3-stage FPN as Figure~\ref{fig:design_choices} (b) suffices, and could also achieve competitive performance compared to those 4-stage backbones (see the above table). Moreover, recent works~\cite{ali2021xcit, li2022exploring} also show that, the plain ViTs can achieve promising performance on downstream tasks with minor modifications in FPN.

\textbf{Downsample with residual patch merging.} To conduct feature downsampling in hierarchical ViTs, two commonly-used modules are stride-2 convolution \cite{wang2021pyramid, wang2022pvt} and linear patch merging in Swin~\cite{liu2021swin}. In this paper, we adopt a patch merging module for better efficiency and a more homogeneous way in the transformer, with an additional cheap residual branch for better gradient flows, as shown in Figure~\ref{fig:design_choices} (c). The following table shows that our residual patch merging achieves slightly higher accuracy with negligible efficiency drop.

\begin{table}[h]
	\renewcommand\arraystretch{1.3}
	\setlength\tabcolsep{1.2mm}
	\centering
	\footnotesize
	\begin{tabular}{l|cccc}
	    Method & Params (M) & FLOPs (G) & Throughput (images/s) & Top-1 (\%) \\
	    \shline
	    w/o residual in patch merging & 9.3 & 0.73 & 2597 & 78.4\\
	    LightViT-T & 9.4 & 0.73 & 2578 & \textbf{78.7}\\
	\end{tabular}
\end{table}

\textbf{Overlapping patch embedding.} Previous methods~\cite{xiao2021early, wang2022pvt} show that, replacing the original patch embedding in plain ViTs with overlapping path embedding (OPE) could benefit the performance and training robustness. In this paper, we also conduct an OPE stem (see Figure~\ref{fig:arch}), and obtain higher performance on ImageNet as in the table below. 

\begin{table}[h]
    \vspace{-2mm}
	\renewcommand\arraystretch{1.3}
	\setlength\tabcolsep{1.2mm}
	\centering
	\footnotesize
	\begin{tabular}{l|cccc}
	    Method & Params (M) & FLOPs (G) & Throughput (images/s) & Top-1 (\%) \\
	    \shline
	    w/o OPE & 9.5 & 0.74 & 2652 & 77.4\\
	    LightViT-T & 9.4 & 0.73 & 2578 & \textbf{78.7}\\
	\end{tabular}
\end{table}

\input{tables/archs.tex}

\subsection{Architecture variants}

We design a series of LightViT models on different scales to validate our effectiveness on light-weight models. The macro structure of our models is illustrated in Figure~\ref{fig:arch}. We first conduct a stem block to embed the input image into image tokens with $\mathrm{stride}=8$, where several convolution layers are included. For the main body of our network, we construct three stages (S1-S3) with the same LightViT blocks inside, and a residual patch merging layer is conducted before S2 and S3 for feature downsampling. The window size $S$ of attention is set to $7$, and reduction ratios $r$ of spatial and channel attentions in FFN are equal to $4$. Detailed settings of our variants are summarized in Table~\ref{tab:archs}.

\section{Experiments}
We validate the efficacy of our proposed model on various vision tasks: image classification, object detection, instance segmentation.

\subsection{Image classification on ImageNet}

\input{tables/strategy.tex}
\textbf{Training strategy.} We train our models on ImageNet-1K dataset~\cite{Imagenet}, and validate the top-1 accuracy on ImageNet validation set. We adopt common data augmentations on ViTs including RandAugmentation~\cite{cubuk2020randaugment}, MixUp~\cite{zhang2017mixup}, \textit{e.t.c}. Detailed training strategy refers to Table~\ref{tab:imgnet_strategy}.

\textbf{Experimental results.} Our performance on ImageNet validation set is summarize in Table~\ref{tab:imagenet}. LightViT outperforms recent efficient ViTs under the basic $224\times224$ resolution, especially having large improvements on light-weight scales (less than 2G FLOPs). For instance, LightViT-T obtains a record $78.7\%$ accuracy with 0.7G FLOPs, significantly outperforms those ViT variants with attention-convolution hybrid blocks. Besides, our model also obtains higher throughput compared to existing efficient ViTs, and achieves better FLOPs-accuracy and latency-accuracy trade-offs, as shown in Figure~\ref{fig:f1}.

\input{tables/main_exp.tex}

\subsection{Object detection and instance segmentation}
\textbf{Training strategy.} We conduct experiments on MS-COCO dataset~\cite{lin2014microsoft} and adopt the Mask R-CNN architecture with an FPN~\cite{lin2017feature} neck for fair comparisons. Since LightViT only has three stage, we make a simple modification to make it compatible with the existing architecture. As shown in Figure~\ref{fig:design_choices}(b), we append a $2\times 2$ transposed convolution with stride 2 to upsample the largest output of the top-down path to the size of $1/4$, forming a pyramid of 4 levels. The models are fine-tuned from the ImageNet pre-trained weights. We use the AdamW~\cite{kingma2014adam} optimizer for training for which the hyper-parameters are: the batch size as 16, the learning rate as 1$\mathrm{e}^{-4}$, weight decay as 0.05, and the stochastic depth~\cite{huang2016deep} ratios as the ones used in the pre-training. Following the common practice, we adopt the same data augmentations scheme as~\cite{liu2021swin} and the $1\times$/$3\times$ training schedule of mmdetection~\cite{chen2019mmdetection}, which has 12/36 training epochs in total and decays the learning rate by a factor of 10 at the $3/4$ and $11/12$ of the total epochs. The standard metrics of the COCO dataset are used here to evaluate the performance, including the Average Precision (AP), AP$_{50}$, and AP$_{75}$ for both object detection and instance segmentation.

\textbf{Experimental results.} We report the results on MS-COCO dataset in Table~\ref{tab:det}. LightViT performs on par with the recent 4-stage ViTs with a similar amount of FLOPs. Specifically, LightViT-S achieves 40.0\% AP${^b}$ and 37.4\% AP${^m}$ using the $1\times$ schedule, which is better than the hybrid method PVT-T with $\sim$200 GFLOPs.
\input{tables/det.tex}

\subsection{Ablation studies}

\textbf{Ablation on proposed aggregation scheme in attention and FFN.} In LightViT, we improve the vanilla local window self-attention with additional global attention for global representations. While in FFN, we propose a bi-dimensional attention module to refine the features for more efficient filtering. Here we conduct experiments for ablations of these components in Table~\ref{tab:ab_components}. \textbf{Local self-attention vs. vanilla global self-attention:} The vanilla self-attention module in ViT~\cite{dosovitskiy2020image} can perform global and dense attentions on tokens. We compare it with the widely-used local-window self-attention. The results show that local-window self-attention has higher accuracy on light-weight model due to better inductive bias. \textbf{+ global attention}: Our proposed global attention gains significant improvements ($76.9\%\sim78.0\%$) over the local self-attention baseline, and only has a minor increase on FLOPs. \textbf{+ spatial attention.} Spatial attention in FFN further achieves an $0.4\%$ higher accuracy with negligible computation cost, as it explicitly captures the spatial dependencies and selectively focus on the salient tokens to capture the image structure better. \textbf{+ channel attention.} Our final architecture with channel attention on LightViT achieves the best 78.7\% accuracy. Compared with our local window self-attention baseline, LightViT-T gains a significant 1.8\% improvement, with better feature aggregation scheme equipped in attention and FFN.

\input{tables/ab_components.tex}

\section{Conclusion}

This paper proposes a new series of light-weight ViTs dubbed LightViT. While most recent works aim to combine convolutions and transformers in efficient ViTs, we seek a better performance-efficiency trade-off on the pure ViT blocks without convolution. This paper proposes a different and more homogeneous way to perform better information aggregation on self-attention and feed-forward networks and achieve better performance on ImageNet classification and object detection than those hybrid models. We hope the work can help future related research study the difference between convolutions and transformers, and look forward to exploring better feature aggregation schemes upon them.

{
\small
\bibliography{main}
\bibliographystyle{abbrv}
}


\section{Appendix}

\subsection{More ablation studies}

\textbf{Number of global tokens.} We investigate the effects of the numbers of global tokens in our model. As shown in Table~\ref{tab:nog}, we train our LightViT-T with $0$, $2$, $4$, $8$, $16$, and $32$ global tokens. We can see that the performance can be improved with only a small number of tokens, \eg, only $2$ tokens can improve the baseline by 0.5\% with 0.01G addition on FLOPs. Besides, the number of $16$ seems to be saturate for the LightViT-T model, as there are no further gains on $32$ tokens. As a result, we set the tokens to a moderate size of $8$ for a better efficiency-accuracy trade-off.

\input{tables/ab_token_numbers.tex}

\subsection{Discussion}
\textbf{Limitations.} LightViT could significantly improve the transformer blocks without leveraging convolutions. However, its inference efficiency on edge devices might be worse than the convolutions, as current inference frameworks on edge devices are better optimized for convolution computations.

\textbf{Society impacts.} Investigating the effects of the proposed model requires large consumptions on computation resources, which can potentially raise the environmental concerns. However, it is valuable for us to explore efficient models, which can save a large volume of computation resources in training and deployment.

\subsection{Implementation of local-global broadcast of attention}
\begin{figure}[h]
    \centering
    \includegraphics[width=\textwidth]{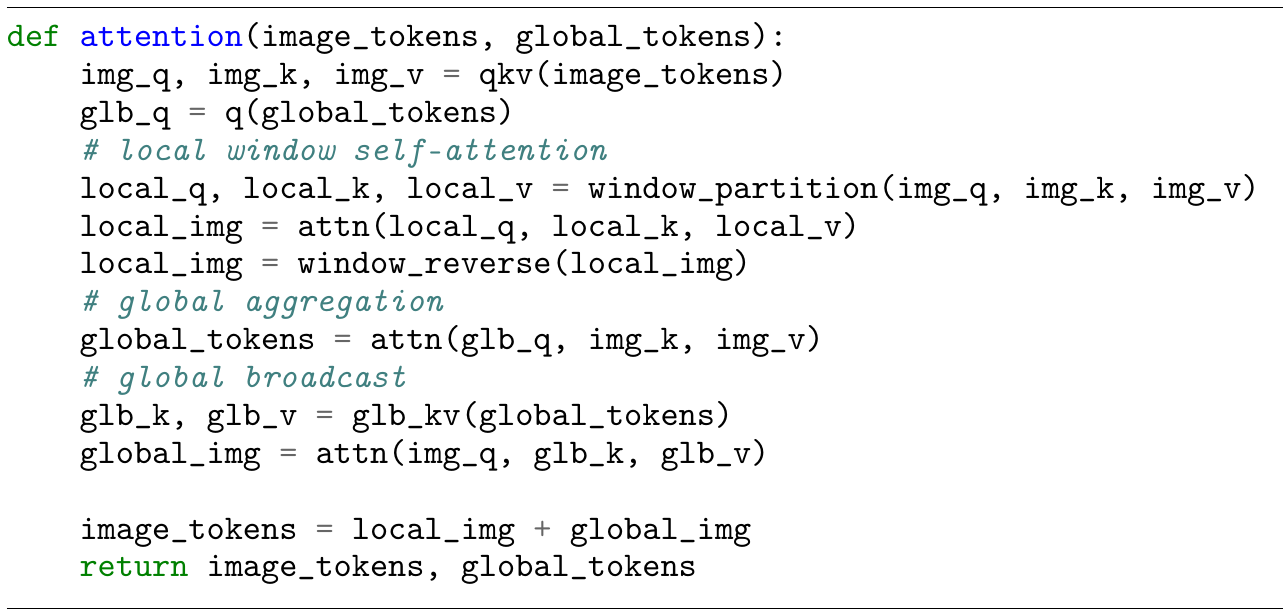}
    \vspace{-2mm}
    \caption{The PyTorch-like pseudo code of our attention module.}
    \label{fig:code_attn}
\end{figure}

The pseudo code of our attention module is shown in Figure~\ref{fig:code_attn}.

\subsection{Implementation of bi-dimensional attention on FFN}
\begin{figure}[h]
    \centering
    \includegraphics[width=\textwidth]{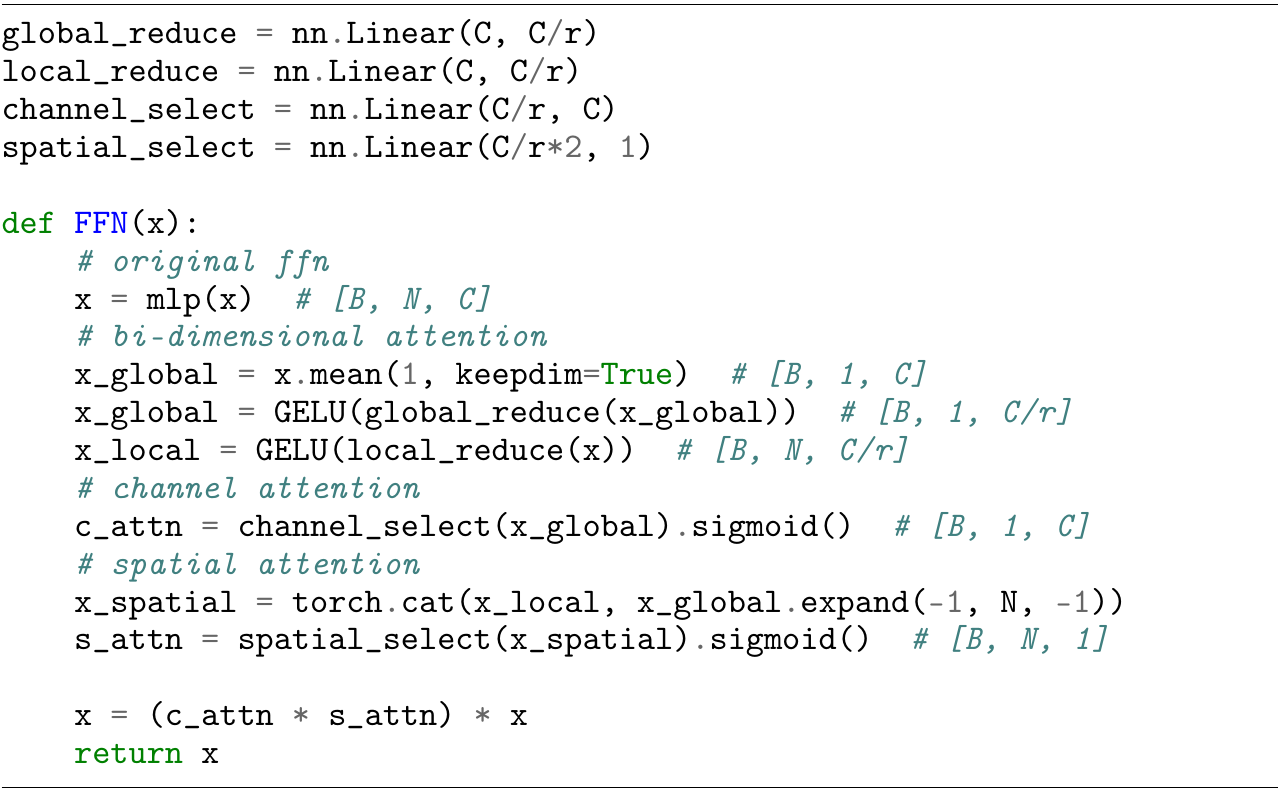}
    \vspace{-2mm}
    \caption{The PyTorch-like pseudo code of our bi-dimensional attention on FFN.}
    \label{fig:code_ffn}
\end{figure}

The pseudo code of our bi-dimensional attention on FFN is shown in Figure~\ref{fig:code_ffn}.

\end{document}

%% file: lib/my_instructions.tex
\input{lib/math_commands.tex}

\definecolor{mycyan}{RGB}{212, 239, 251}
\usepackage{colortbl}
\usepackage{xcolor}
\definecolor{mygray}{gray}{.9}
\definecolor{goldenrod}{RGB}{245,245,220}
\newlength\savewidth\newcommand\shline{\noalign{\global\savewidth\arrayrulewidth\global\arrayrulewidth 1pt}\hline\noalign{\global\arrayrulewidth\savewidth}}
\newcolumntype{a}{>{\columncolor{mygray}}c}
\usepackage{fontawesome5}
\usepackage{booktabs}
\usepackage{makecell}
\usepackage{fontawesome5}
\usepackage{lipsum}
\usepackage{comment}
\usepackage{multirow}
\usepackage{wrapfig}
\usepackage{algorithm}
\usepackage{algorithmic}
\usepackage{xfrac}  

\usepackage{graphicx}
\usepackage{color}

\usepackage[english]{babel}
\usepackage{amsthm}
\usepackage{subfig}
\definecolor{darkgreen}{rgb}{0,0.7,0}

\definecolor{mygraytext}{gray}{.75}
\newcommand{\graytext}[1]{{\color{mygraytext}{#1}}}

\def\eg{\emph{e.g.}} 
\def\ie{\emph{i.e.}}


%% file: lib/math_commands.tex
\usepackage{amsmath,amsfonts,bm}

\def\eqref#1{equation~\ref{#1}}

\def\1{\bm{1}}

\DeclareMathAlphabet{\mathsfit}{\encodingdefault}{\sfdefault}{m}{sl}
\SetMathAlphabet{\mathsfit}{bold}{\encodingdefault}{\sfdefault}{bx}{n}



%% file: tables/archs.tex
\begin{table}[h]
	\renewcommand\arraystretch{1.4}
	\setlength\tabcolsep{3mm}
	\centering
	\caption{\textbf{LightViT architecture variants.} \texttt{B}: number of blocks. \texttt{C}: number of channels. \texttt{H}: number of heads. We set numbers of global tokens to $[8, 16, 24]$ (T, S, B).}
	\label{tab:archs}
	\footnotesize
	\begin{tabular}{l|c|c|c|c}
	    \shline
	    Stage & Stride & LightViT-T & LightViT-S & LightViT-B\\
	    \shline
	    S0: Stem & $\sfrac{1}{8}$ & \texttt{C=64\ } & \texttt{C=96\ \ } & \texttt{C=128}\\
        S1: LightViT-Block & $\sfrac{1}{8}$ & \texttt{B=2 C=64\ \ H=2} & \texttt{B=2 C=96\ \ H=3\ } & \texttt{B=3\ C=128 H=4\ }\\
        S2: LightViT-Block & $\sfrac{1}{16}$ & \texttt{B=6 C=128 H=4} & \texttt{B=6 C=192 H=6\ } & \texttt{B=8 C=256 H=8\ }\\
        S3: LightViT-Block & $\sfrac{1}{32}$ & \texttt{B=6 C=256 H=8} & \texttt{B=6 C=384 H=12} & \texttt{\ B=6\ C=512 H=16}\\
	    \shline
	\end{tabular}
	\vspace{-2mm}
\end{table}

%% file: tables/strategy.tex
\begin{wraptable}{r}{0.4\textwidth}
	\renewcommand\arraystretch{1.1}
	\setlength\tabcolsep{2mm}
	\centering
	\vspace{-14mm}
	\caption{\textbf{Training settings on ImageNet dataset.}}
	\label{tab:imgnet_strategy}
	\footnotesize
	\begin{tabular}{l|l}
	    \shline
	    Config & Value\\
	    \shline
	    Batch size & 1024 \\
	    Optimizer & AdamW\\
	    Weight decay & 0.04\\
	    \hline
	    LR decay & cosine\\
	    Base LR & 1e-3\\
	    Minimum LR & 1e-6\\
	    Warmup LR & 1e-7\\
	    Warmup epochs & 20\\
	    Training epochs & 300\\
	    \hline
	    Augmentation & RandAug (2, 9)~\cite{cubuk2020randaugment} \\
	    Color jitter & 0.3\\
	    Mixup alpha & 0.2\\
	    Cutmix alpha & 1.0\\
	    Erasing prob. & 0.25\\
	    Drop path rate & 0.1 (T, S), 0.3 (B)\\
	    \shline
	\end{tabular}
	\vspace{-6mm}
\end{wraptable}

%% file: tables/main_exp.tex
\begin{table}[t]
	\renewcommand\arraystretch{1.4}
	\setlength\tabcolsep{3mm}
	\centering
	\caption{\textbf{Image classification performance on ImageNet validation dataset.} Throughput is measured on a single V100 GPU following \cite{touvron2021training, liu2021swin}. \textit{Hybrid} denotes using both attention and convolution in blocks. All models are trained and evaluated on $224\times224$ resolution. $\dagger$: accuracy reported by DeiT~\cite{touvron2021training}.}
	\label{tab:imagenet}
	\footnotesize
	\begin{tabular}{l|ccccc}
	    \shline
	    Model & \thead{Block\\type} &\thead{Params\\(M)} & \thead{FLOPs\\(G)} & \thead{Throughput\\(image/s)} & \thead{Top-1\\(\%)} \\
	    \shline
	    RegNetY-800M~\cite{radosavovic2020designing} & CNN & 6.3 & 0.8 & 3321 & 76.3\\
	    PVTv2-B0~\cite{wang2022pvt} & Hybrid & 3.4 & 0.6 & 2324 & 70.5 \\
	    SimViT-Micro~\cite{li2021simvit} & Hybrid & 3.3 & 0.7 & 1004 & 71.1 \\
	    MobileViT-XS~\cite{mehta2021mobilevit} & Hybrid & 2.3 & 0.7 & 1581 & 74.8\\
	    LVT~\cite{yang2021lite} & Hybrid & 5.5 & 0.9 & 1545 & 74.8\\
	    \rowcolor{goldenrod}
	    LightViT-T & Transformer & 9.4 & 0.7 & 2578 & \textbf{78.7}\\
	    \hline
	    RegNetY-1.6G~\cite{radosavovic2020designing} & CNN & 11.2 & 1.6 & 1845 & 78.0\\
	    MobileViT-S~\cite{mehta2021mobilevit} & Hybrid & 5.6 & 1.1 & 1219 & 78.4\\
	    PVTv2-B1~\cite{wang2022pvt} & Hybrid & 13.1 & 2.1 & 1231 & 78.7\\
	    ResT-Small~\cite{zhang2021rest} & Hybrid & 13.7 & 1.9 & 1298 & 79.6\\
	    DeiT-Ti~\cite{touvron2021training} & Transformer & 5.7 & 1.3 & 2612 & 72.2 \\
	    \rowcolor{goldenrod}
	    LightViT-S & Transformer & 19.2 & 1.7 & 1467 & \textbf{80.8}\\
	    \hline
	    RegNetY-4G$^\dagger$~\cite{radosavovic2020designing} & CNN & 21.0 & 4.0 & 1045 & 80.0\\
	    Twins-PCPVT-S~\cite{chu2021twins} & Hybrid & 24.1 & 3.8 & 807 & 81.2\\
	    ResT-Base~\cite{zhang2021rest} & Hybrid & 30.3 & 4.3 & 735 & 81.6\\
	    PVTv2-B2~\cite{wang2022pvt} & Hybrid & 25.4 & 4.0 & 695 & 82.0\\
	    DeiT-S~\cite{touvron2021training} & Transformer & 22 & 4.6 & 961 & 79.8\\
	    Swin-T~\cite{liu2021swin} & Transformer & 29 & 4.9 & 765 & 81.3\\
	    \rowcolor{goldenrod}
	    LightViT-B & Transformer & 35.2 & 3.9 & 827 & \textbf{82.1}\\
	    \shline
	\end{tabular}
	\vspace{-2mm}
\end{table}

%% file: tables/det.tex
\begin{table}[h]
	\renewcommand\arraystretch{1.4}
	\setlength\tabcolsep{0.5mm}
	\centering
	\caption{\textbf{Object detection and instance segmentation performance on COCO \texttt{val2017}.} The FLOPs are measured on $800\times1280$. All the models are pretrained on ImageNet-1K.}
	\label{tab:det}
	\footnotesize
	\begin{tabular}{l|cc|cccccc|cccccc}
	    \shline
	    \multirow{2}*{Backbone} & Params & FLOPs & \multicolumn{6}{c|}{Mask R-CNN 1x schedule} & \multicolumn{6}{c}{Mask R-CNN 3x + MS schedule} \\
	    \cline{4-15}
	    ~ & (M) & (G) & AP${}^{b}$ & AP${}_{50}^{b}$ & AP${}_{75}^{b}$ & AP${}^{m}$ & AP${}_{50}^{m}$ & AP${}_{75}^{m}$ & AP${}^{b}$ & AP${}_{50}^{b}$ & AP${}_{75}^{b}$ & AP${}^{m}$ & AP${}_{50}^{m}$ & AP${}_{75}^{m}$\\
	    \shline
        ResNet-18~\cite{he2016deep} & 31 & 207 & 34.0 & 54.0 & 36.7 & 31.2 & 51.0 & 32.7 & 36.9 & 57.1 & 40.0 & 33.6 & 53.9 & 35.7\\
        ResNet-50~\cite{he2016deep} & 44 & 260 & 38.0 & 58.6 & 41.4 & 34.4 & 55.1 & 36.7 & 41.0 & 61.7 & 44.9 & 37.1 & 58.4 & 40.1\\
        ResNet-101~\cite{he2016deep} & 101 & 493 & 40.4 & 61.1 & 44.2 & 36.4 & 57.7 & 38.8 & 42.8 & 63.2 & 47.1 & 38.5 & 60.1 & 41.3\\
        \hline
        PVT-T~\cite{wang2021pyramid} & 33 & 208 & 36.7 & 59.2 & 39.3 & 35.1 & 56.7 & 37.3 & 39.8 & 62.2 & 43.0 & 37.4 & 59.3 & 39.9\\
        PVT-S~\cite{wang2021pyramid} & 44 & 245 & 40.4 & 62.9 & 43.8 & 37.8 & 60.1 & 40.3 & 43.0 & 65.3 & 46.9 & 39.9 & 62.5 & 42.8\\
        PVT-M~\cite{wang2021pyramid} & 64 & 302 & 42.0 & 64.4 & 45.6 & 39.0 & 61.6 & 42.1 & 44.2 & 66.0 & 48.2 & 40.5 & 63.1 & 43.5\\
        \hline
        LightViT-T & 28 & 187 & 37.8 & 60.7 & 40.4 & 35.9 & 57.8 & 38.0 & 41.5 & 64.4 & 45.1 & 38.4 & 61.2 & 40.8\\
        LightViT-S & 38 & 204 & 40.0 & 62.9 & 42.6 & 37.4 & 60.0 & 39.3 & 43.2 & 66.0 & 47.4 & 39.9 & 63.0 & 42.7\\
	    LightViT-B & 54 & 240 & 41.7 & 64.5 & 45.1 & 38.8 & 61.4 & 41.4 & 45.0 & 67.9 & 48.8 & 41.2 & 64.8 & 44.2\\
	    \shline
	\end{tabular}
	\vspace{4mm}
\end{table}

%% file: tables/ab_components.tex
\begin{table}[h]
	\renewcommand\arraystretch{1.4}
	\setlength\tabcolsep{2mm}
	\centering
	\caption{\textbf{Ablation of the components of our architecture with LightViT-T settings.}}
	\label{tab:ab_components}
	\footnotesize
	\begin{tabular}{cc|cc|ccc}
	    \shline
	    \multicolumn{2}{c|}{Attention} & \multicolumn{2}{c|}{FFN} & Params & FLOPs & Top-1 \\
	    local self-attn. & global attn. & spatial attn. & channel attn. & (M) & (G) & (\%) \\
	    \shline
	    \graytext{\faTimes} & \graytext{\faTimes} & \graytext{\faTimes} & \graytext{\faTimes} & 8.0 & 0.88 & 76.5\\
	    \faCheck & \graytext{\faTimes} & \graytext{\faTimes} & \graytext{\faTimes} & 8.0 & 0.66 & 76.9\\
	    \graytext{\faCheck} & \faCheck & \graytext{\faTimes} & \graytext{\faTimes} & 8.8 & 0.71 & 78.0\\
	    \hline
	    \graytext{\faCheck} & \graytext{\faCheck} & \faCheck & \graytext{\faTimes} & 9.1 & 0.73 & 78.4\\
	    \rowcolor{goldenrod}
	    \graytext{\faCheck} & \graytext{\faCheck} & \graytext{\faCheck} & \faCheck & 9.4 & 0.73 & \textbf{78.7}\\
	    \shline
	\end{tabular}
	\vspace{4mm}
\end{table}

%% file: tables/ab_token_numbers.tex
\begin{table}[h]
	\renewcommand\arraystretch{1.3}
	\setlength\tabcolsep{3mm}
	\centering
	\footnotesize
	\caption{\textbf{Ablation of numbers of global tokens.}} 
	\label{tab:nog}
	\begin{tabular}{c|cc}
	    \shline
	    \#global tokens & FLOPs (G) & Top-1 (\%) \\
	    \shline
	    0 & 0.68 & 77.3\\
	    2 & 0.69 & 77.8\\
	    4 & 0.70 & 78.3\\
	    \rowcolor{goldenrod}
	    8 & 0.73 & 78.7\\
	    16 & 0.80 & 78.8\\
	    32 & 0.92 & 78.8\\
	    \shline
	\end{tabular}
\end{table}